\documentclass[preprint,12pt,number]{elsarticle}

\usepackage{cite}
\usepackage{amsmath,amssymb,amsfonts}
\usepackage{lineno}
\usepackage{algorithm}
\usepackage{algpseudocode}
\usepackage{subfigure}
\usepackage{textcomp}
\usepackage[utf8]{inputenc} 
\usepackage{graphicx} 
\usepackage[parfill]{parskip} 
\usepackage{booktabs} 
\usepackage{array} 
\usepackage{paralist} 
\usepackage{verbatim} 
\usepackage{subfig} 
\usepackage{xcolor}
\usepackage{threeparttablex}
\usepackage{longtable}
\usepackage{subcaption}
\usepackage{multirow}
\usepackage{tabularx}
\usepackage{threeparttable}
\usepackage{arydshln} 

\newcolumntype{L}[1]{>{\raggedright\arraybackslash}p{#1}} 

\title{Alliance Beats Isolation: Unifying Heterogeneous Allied Datasets Improves Classifier Performance}

\journal{Data and Knowledge Engineering}

\begin{document}

\begin{frontmatter}
\title{} 
\author[1]{Girish Keshav Palshikar\corref{cor1}} 
\affiliation[1]{organization={Cummins College of Engineering for Women},
            addressline={Karvenagar}, 
            city={Pune},
            postcode={411052}, 
            state={Maharashtra},
            country={India}}
\cortext[cor1]{Corresponding author}

\begin{abstract}
In many application domains, such as student dropout, insurance fraud, loan approval, and machine failures, several labelled public datasets are available where (i) data is about the same type of objects but the set of actual underlying objects are disjoint; and (ii) the class labels are same; and (iii) the feature spaces of the datasets are largely distinct (heterogeneous), with a few shared features. We call such datasets as {\em allied}. A single classifier cannot be trained on both datasets together, and one classifier trained on one dataset cannot be tested on the other. In this paper, we propose a method to merge the feature-spaces into a single feature-space for a pair of given allied heterogeneous datasets. We then use a matrix completion method to create a unified dataset based on the merged feature-space. The hypothesis is that the merged representation facilitates the transfer of classification knowledge from one dataset to another. We conduct experiments on several pairs of allied, heterogeneous datasets and several classifiers to demonstrate that any classifier trained on the unified representation always outperforms classifiers separately trained on the constituent allied datasets on several pairs of allied datasets. This work provides an easy way to substantially improve classifier performance by unifying and using multiple allied datasets together. 
\end{abstract}

\begin{graphicalabstract}
\includegraphics[width=\textwidth,height=7cm]{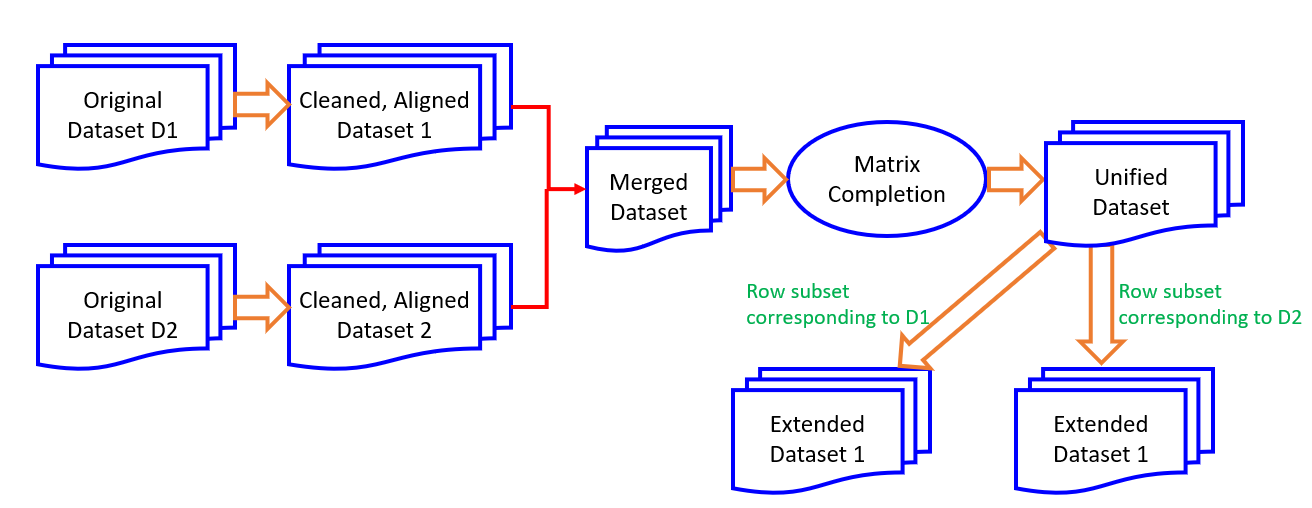}
\end{graphicalabstract}

\begin{highlights}
\item Defines the notion of allied datasets, having data about similar objects, similar class-labels and overlapping but otherwise heterogeneous feature-spaces
\item Proposes a method to merge the feature-spaces into a single feature-space for a pair of allied datasets
\item Uses a matrix completion method to create a unified dataset based on the merged feature-space 
\item Hypothesizes that merged representation facilitates transfer of classification knowledge from one dataset to another
\item Experimentally demonstrates that any classifier trained on the unified representation always outperforms classifiers separately trained on the constituent allied datasets on several pairs of allied datasets
\end{highlights}


\end{frontmatter}


\maketitle

\section{Introduction}\label{sec1}

In many domains, several labelled public datasets are available with the following characteristics: (i) data is about the same type of objects but the set of actual underlying objects are disjoint; and (ii) the class labels are same; and (iii) the feature spaces of the datasets are largely distinct, with a few shared features. We call such datasets as {\em allied}. For simplicity, we assume that the class labels are binary $\{0, 1\}$, with $1$ denoting the minority class of interest (e.g., a dropped out student, or a fraudulent claim). Generalization to more than two class labels is simple. For simplicity, we assume that we are given two allied, heterogeneous datasets, denoted $D_1$ and $D_2$, each as a single table. Generalization to more than two allied constituent datasets is simple and discussed later. If the dataset is split across multiple tables, we assume it can be brought into a single table form, say by denormalization.

\small
\begin{table}
\begin{threeparttable}
\caption{Examples of Allied Datasets} \label{tab_ex}
\centering
\begin{tabular}{|L{6cm}||l|l|l|l|l|}
\hline
\textbf{Allied Datasets} & \textbf{\#Rows} & \textbf{\#Shared} & \textbf{Minority} & \textbf{\#Base}\\
                          &  \textbf{\#Cols}  & \textbf{Cols}      & \textbf{Class Label \%} & \textbf{line $\mathbf{F_1}$}\tnote{0}\\
\hline\hline
Open University Learning Analytics Dataset (OULAD) & $32593,27$ & \multirow{2}{*}{$7$} & $31.1$\%:{\footnotesize {\sf DROPOUT}} & $0.7749$\\
\hdashline
Higher Education Predictors of Student Retention (HEPSR)\tnote{1} & $3630,35$ & & $26.7$\%:{\footnotesize {\sf DROPOUT}} & $0.8592$\\
\hline\hline
Vehicle Insurance Claim Fraud Detection (Fraud Oracle)\tnote{2} & $15420,30$ & \multirow{2}{*}{$13$} & $5.99$\%:{\footnotesize {\sf FRAUD}} & $0.0889$\\
\hdashline
Automobile Insurance Claims (Mendeley)\tnote{3} & $1000,34$ &  & $24.7$\%:{\footnotesize {\sf FRAUD}} & $0.4497$\\
\hline\hline
Loan Approval (Umair)\tnote{4} & $30000,38$ & \multirow{2}{*}{$10$}  & $42.26$\%:{\footnotesize {\sf REJECTED}} & $0.1000$\\
\hdashline
Loan Approval (Archit Sharma)\tnote{5} & $4269,13$ & & $37.78$\%:{\footnotesize {\sf REJECTED}} & $0.9782$\\
\hline\hline
Machine Failure (Saquib)\tnote{6} & $3000,6$ & \multirow{2}{*}{$6$}  & $9.93$\%:{\footnotesize {\sf FAILED}} & $0.0074$\\
\hdashline
Machine Failure (IIoT)\tnote{7} & 1000,16 & & $16.2$\%:{\footnotesize {\sf FAILED}} & $0.0743$\\
\hline\hline
\end{tabular}
\begin{tablenotes}
\item[0] {\footnotesize All values in this table are based on pre-processed dataset versions aligned with each other}
\item[1] {\footnotesize Rows having class label {\footnotesize {\sf Enrolled}} were removed}
\item[2] {\footnotesize kaggle.com/datasets/shivamb/vehicle-claim-fraud-detection}
\item[3] {\footnotesize https://data.mendeley.com/datasets/992mh7dk9y/2}
\item[4] {\footnotesize kaggle.com/datasets/mmumairkhattak/loan-approval-dataset-2026-credit-risk-and-bank-a}
\item[5] {\footnotesize kaggle.com/datasets/mytalkwithyou/bank-loan-approval-dataset}
\item[6] {\footnotesize kaggle.com/datasets/saquib7hussain/machine-failure-prediction-dataset?select=machine\_failure\_data.csv}
\item[7] {\footnotesize kaggle.com/datasets/zara2099/iiot-sensor-data-for-predictive-maintenance}
\end{tablenotes}
\end{threeparttable}
\end{table}
\normalsize

Table~\ref{tab_ex} shows some allied public datasets, each with two {\em constituent datasets}. One example has two constituent datasets about student attrition (OULAD \citep{KHZ17} and HEPSR \citep{MTMB21}), where the objects are students and the courses they are enrolled in and the class label is about whether the student completed the course or dropped out of it. The individual students in the two datasets are unrelated and disjoint. Both datasets share some common features (with possibly different names) but the other features are specific to each dataset. These datasets are inherently {\em heterogeneous} i.e., they differ in terms of feature spaces, schema representations, and data distributions. Another example shows two constituent datasets about automobile insurance claims, where the main objects are claims, insurance policies, vehicles and policy-holders, and the class label indicates whether the claim was fraudulent or not. The claims in the two datasets are unrelated (different). The feature spaces are heterogeneous. 

Such allied labelled datasets in a domain are collected independently by different teams and their data elements (e.g., rows) refer to {\em different, unrelated} objects. Because of these differences, such allied  datasets cannot be directly merged to create a larger, {\em unified} dataset for training. A common way to use these allied datasets in ML is to train and evaluate {\em different} prediction models, one for each of the allied datasets. Consequently, each such model is tailored only to the characteristics of a dataset. Each dataset-specific model discards the non-common attributes, and completely ignores the challenges posed by the heterogeneous feature spaces and different data distributions. Such approaches lead to information loss and fail to exploit the complementary knowledge available across multiple datasets. The models exhibit limited generalizability when it is desired to be applied to other allied datasets. In fact, a model trained on, say $D_1$ cannot even be directly tested on $D_2$ because of the presence of non-shared attributes in $D_2$ which were not available when the model was trained on $D_1$ . 

The research questions addressed in this paper are: 
\begin{enumerate}
\item \textbf{Super-Dataset (Dataset Unification)}: How can $D_1$ and $D_2$ be {\em unified} ({\em synthesized}) into a {\em super-dataset} $D_{12}$, which effectively combines the information (and knowledge) hidden in the two constituent datasets $D_1$ and $D_2$ and which minimizes the information loss? Any proposed data unification method should handle disjoint underlying objects, feature-space heterogeneity, data distribution differences, and domain differences. What would be the attributes in $D_{12}$? How to construct the rows in $D_{12}$ from rows in $D_1$ and $D_2$? 
\item Would any standard predictive ML model trained on the unified dataset $D_{12}$ be more accurate than individual models (of the same type) $M_1$ and $M_2$ trained on $D_1$ and $D_2$ respectively? The model should be capable of providing robust and transferable predictions across multiple data contexts. 
\end{enumerate}
We answer both questions affirmatively in this paper. 

There is limited research onconstructing a unified super-dataset from multiple allied datasets while preserving both shared and dataset-specific information. As a result, it is not clear how a generalized ML model can learn from such allied datasets and accurately make predictions for objects already within either of the constituent datasets, or for unseen objects that may come in the format of data in $D_1$ or in $D_2$. Therefore, there is a need of developing an integrated framework that allows the user to building super-classifiers from allied datasets. This integrated framework would enable the development of a scalable, institution-independent unified datasets and associated classification (predictive) models. The idea is to leverage the diversity of heterogeneous allied data sources to improve prediction accuracy, robustness, and generalizability.

This paper is organized as follows. Section~\ref{sec2} formalizes the problem of unification of allied datasets. Section~\ref{sec_related} discusses relevant research work.  Section~\ref{sec_uni} gives details of the data unification procedure which can use any suitable matrix completion method. Section~\ref{sec_results} gives experimental results and provides some insights derived from them. Section~\ref{sec_last} provides conclusions and discusses some future work. 

\section{Problem Formalization}\label{sec2}

\begin{figure}[htbp]
\includegraphics[width=0.9\columnwidth]{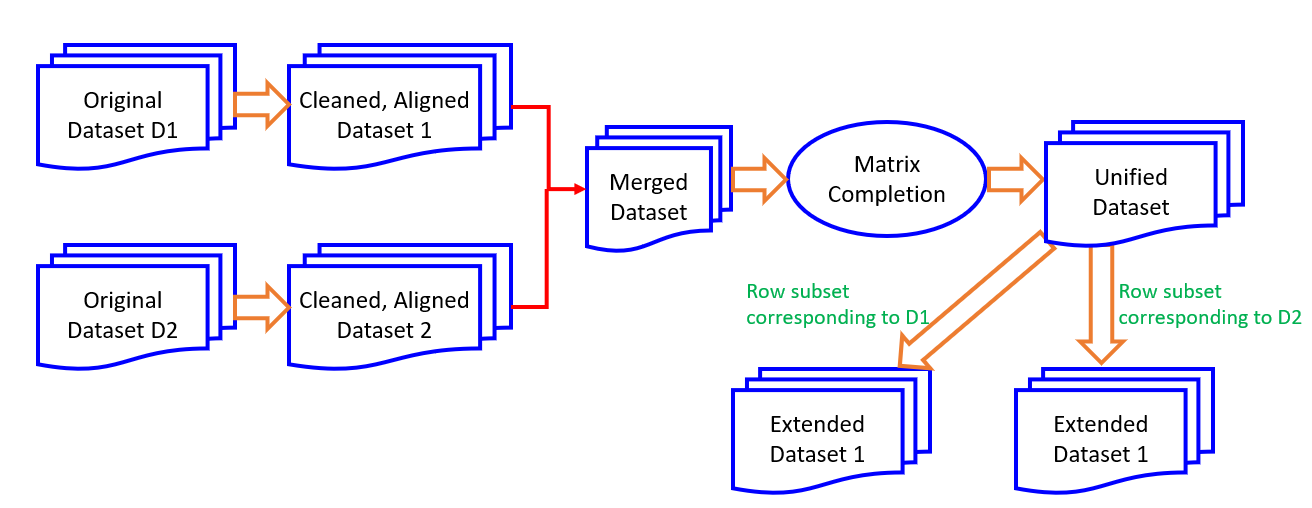}
\caption{Steps in creating a unified dataset from constituent allied datasets.}
\label{fig_data_unification}
\end{figure}

Let $D_1$ and $D_2$ be two given aligned datasets, both having data about the same type of objects, having disjoint sets of underlying objects, having some shared (common) features and otherwise heterogeneous features, and having  the same binary class labels. Fig.~\ref{fig_data_unification} shows the steps involved in creating a unified representation from constituent datasets $D_1$ and $D_2$. In the first step, we {\em align} (or {\em clean}) these two given allied datasets:
\begin{enumerate}
\item Ensure all columns shared between the two datasets have the same names (change names, if required). Ensure that each shared attribute has exactly the same meaning and similar values in both datasets; e.g., consistently mapping value in column {\small {\sf Gender}} in both datasets, or converting {\small {\sf DOB}} to {\small {\sf Age}}).
\item Remove columns that are the primary keys for the table (e.g., {\small {\sf Student\_ID}}, or {\small {\sf Claim\_ID}}).
\item Replace values in every categorical column with numbers.
\item Ensure each shared categorical column has the same set of possible values. 
\item Ensure the class labels are binary $\{0, 1\}$ in both tables, with $1$ denoting the minority class of interest (e.g., a dropped out student, or a fradulent claim). 
\item Move all shared columns to the beginning of the table, with the shared class label column as the first. 
\item If required, add simple {\em derived attributes} if they can be shared between the datasets.  
\item Remove columns which are unsuitable for any simple conversion to numbers, e.g., {\small {\sf Address}}. 
\item Remove redundant columns; e.g., remove  {\small {\sf DOB}} if a new converted column {\small {\sf Age}} is added. 
\end{enumerate} 

Formally, let $\mathbf{D}_{allied} = \{D_1, D_2\}$ denote a collection of (for simplicity) two given allied and aligned (cleaned) datasets $D_1$ and $D_2$. We further assume that each dataset is a single table, where each row refers to a specific type of object, such as a student or an insurance claim. In general, the two sets of underlying objects for $D_1$ and $D_2$ are completely disjoint. Let the two datasets have $k$ common, shared attributes $\mathbf{C} = \{C_1, C_2, \ldots, C_k\}$ having the same meaning in both.  Let $\mathbf{A_1} = \{C_1, C_2, \ldots, C_k, A^{(1)}_1,  A^{(1)}_2, \ldots, A^{(1)}_{n_1}\}$ and $\mathbf{A_2} = \{C_1, C_2, \ldots, C_k, A^{(2)}_1,  A^{(2)}_2, \ldots, A^{(2)}_{n_2}\}$ denote the sets of attributes of these two datasets ($n_1 + k$ and $n_2 + k$ attributes respectively), each including the common attributes.  

The allied {\small {\sf student\_dropout}} datasets OULAD and HEPSR are about completely disjoint sets of students and only a few features are shared between them (Table~\ref{tab_common}). Similar is the case for the two allied datasets on automobile insurance claims. In HEPSR dataset, we have used the feature names {\small {\sf C1}} and {\small {\sf C2}} to refer to curricular units that the student was approved for enrollment in the first and second semester respectively. The feature {\small {\sf imd\_band}} in OULAD is comparable to the combined value of 3 features in HEPSR. Thus, some pre-processing and standardization is often necessary to make the values in these common features comparable across the two allied datasets.  

\begin{table}[htbp]
\caption{Common Attributes in Two Allied Datasets.}
\begin{center}
\footnotesize
\begin{tabular}{|l||l|}
\hline\hline
\textbf{OULAD} & \textbf{HEPSR} \\
\hline
{\footnotesize {\sf code\_module + code\_presentation}} & {\footnotesize  {\sf Course}}\\
{\footnotesize  {\sf highest\_education}} & {\footnotesize {\sf Previous\_qualification}}\\
{\footnotesize  {\sf gender}} & {\footnotesize  {\sf Gender}}\\
{\footnotesize  {\sf age\_band}} & {\footnotesize  {\sf Age\_at\_enrollment}}\\
{\footnotesize  {\sf disability}} & {\footnotesize  {\sf Educational\_special\_needs}}\\
{\footnotesize  {\sf studied\_credits}} & {\footnotesize  {\sf C1 + C2}}\\
{\footnotesize  {\sf final\_result}} & {\footnotesize  {\sf Target}}\\
{\footnotesize  {\sf imd\_band}} & {\footnotesize {\sf Debtor + Displaced +}}\\
                                      &  {\footnotesize {\sf (1 - Tuition\_fees\_up\_to\_date)}}\\
\hline
\end{tabular}
\label{tab_common}
\end{center}
\end{table}
\normalsize

\section{Related Work}\label{sec_related}

Data integration is a well-known problem in data management, whose aim is to provide simple, coherent, perspective-specific views of large and diverse data sources. It is often formulated as mapping queries between a global (mediated) schema and various source schemas. Techniques such as {\em Extract, Transform, Load (ETL)} and technologies such as {\em Data Warehousing}~ \citep{Inmo05} and {\em Data Lakes}\footnote{https://www.pwc.com/us/en/technology-forecast/2014/cloud-computing/assets/pdf/pwc-technology-forecast-data-lakes.pdf} are widely used for these tasks. In contrast, the aim in this paper is to unify allied datasets for improving classfication. The task in Data imputation~ \citep{Ende10} is to estimate the missing values in given data. This paper has proposed a novel application of the data imputation techniques of matrix completion for the purpose of data unification. 

Many methods have been devised for {\em matrix completion}~ \citep{PV25},  \citep{Jafa22},  \citep{RYLL18}. Matrix completion involves reconstructing a matrix from only a small subset of its observed entries. This problem is important in many applications - e.g., recommender systems, collaborative filtering, and compressed sensing - where collecting all entries of the data matrix is usually too costly. Matrix completion methods are broadly divided into two types: passive and adaptive. Passive methods are widely used in machine learning. Many of these methods assume that the unknown matrix has a low-rank structure, and then pose an optimization problem to find the lowest rank matrix which matches with the observed entries in the given incomplete matrix. {\em Nuclear norm minimization} methods are a convex version of rank minimization and use the sum of singular values to identify low-rank solutions. {\em Alternating minimization} methods factorize the unknown target matrix into a product of two smaller low-rank matrices $X = UV^T$ and alternate the optimization steps between them. {\em Iterative principal component analysis (PCA)} methods fills missing entries with initial guesses, run PCA approximations, update the cells, and repeat until convergence.

{\em Federated Learning}~ \citep{KMBA21} is another related line of research, where broadly the task is privacy-preserving, decentralized ML, involving issues of local training, model distribution, update sharing and aggregating to construct the global ML model. Related techniques of $k$-anonymity, and differential privacy in {\em Privacy-Preserving Data Mining}~ \citep{AY08} aim to learn useful patterns in data while protecting identification and leakage of sensitive information. In this paper, we are not attempting distributed ML nor privacy preservation. 

Finally, {\em co-training}~ \citep{BM98} is a closely related semi-supervised ML algorithm, which splits given lalelled data into two feature-spaces, learns separate classifiers on each view, and then uses their predictions on given unlabelled data to automatically create more labelled data. Crucial assumptions in co-training are (i) that the two feature sets of each instance are conditionally independent given the class-label; (ii) each view is sufficient to predict the class accurately. In this paper,  we are not proposing a new ML algorithm, and we are also not learning two separate classifiers on two datasets. We also do not need any unlabelled data. Our goal is to systematically unify the datasets to improve classification accuracy of {\em any} classifier. We are also not making any independence assumptions between the two feature-spaces; in fact, we have assumed that the two feature-spaces share at least a few features. 

\section{Unifying Allied Datasets into a Super-Dataset}\label{sec_uni}

\subsection{Merged Data Matrix}

\small
\begin{figure*}[htbp]
\[
M = 
\left[
  \begin{array}{ccc|ccc|ccc}
    D^{(1)}_{1,1} & \ldots & D^{(1)}_{1,k} & D^{(1)}_{1,k+1} & \ldots & D^{(1)}_{1,k+n_1} & NA & \ldots & NA\\
    \ldots & \ldots & \ldots & \ldots & \ldots & \ldots & \ldots & \ldots & \ldots\\
    D^{(1)}_{N_1,1} & \ldots & D^{(1)}_{N_1,k} & D^{(1)}_{N_1,k+1} & \ldots & D^{(1)}_{N_1,k+n_1} & NA & \ldots & NA\\
    \hline
    D^{(2)}_{1,1} & \ldots & D^{(2)}_{1,k} & NA & \ldots & NA & D^{(2)}_{1,k+1} & \ldots & D^{(2)}_{1,k+n_2}\\
    \ldots & \ldots & \ldots & \ldots & \ldots & \ldots & \ldots & \ldots & \ldots\\
    D^{(2)}_{N_2,1} & \ldots & D^{(2)}_{N_2,k} & NA & \ldots & NA & D^{(2)}_{N_2,k+1} & \ldots & D^{(2)}_{N_2,k+n_2}\\
  \end{array}
  \right]
\]
\caption{Merged data matrix representation for two aligned (cleaned) allied datasets $D_1$ and $D_2$.}
\label{fig_merged_matrix}
\end{figure*}
\normalsize

The first step to unify the two allied, and aligned datasets $D_1$ and $D_2$ is to construct a merged data matrix $M$ (Fig.~\ref{fig_merged_matrix}), as follows. 
\begin{enumerate}
\item Construct a combined feature set $\mathbf{A}_{comb} = C \cup A^{(1)}_{nonshared} \cup A^{(2)}_{nonshared}$, where $\mathbf{C}$ is the shared common attributes and $\mathbf{A}^{(1)}_{nonshared} = \{A^{(1)}_1,  A^{(1)}_2, \ldots, A^{(1)}_{n_1}\}$ is the set of features present in $D_1$ but not in $D_2$.  $\mathbf{A}^{(2)}_{nonshared}$ is similarly defined. 
\item Construct a combined (merged) data matrix $M$ with $N_1 + N_2$ rows and $|A_{comb}| = k + n_1 + n_2$ columns, where $N_1$ and $N_2$ are the number of rows in $D_1$ and $D_2$ respectively. For the $i$-th row, $1 \leq i \leq N_1$, their first $k + n_1$ columns are filled with the corresponding values in the $i$-th row in $D_1$ and the next $n_2$ columns are filled with $NA$ values. Then for the $N_1 + j$-th row, $1 \leq j \leq N_2$, their first $k$ columns are filled with the corresponding shared columns in $D_2$, next $n_1$ columns are filled with $NA$ values, and the last $n_2$ columns are filled with the corresponding values in the $j$-th row in $D_2$. Here, $NA$ indicates a missing value. We have assumed that in both $D_1$ and $D_2$ the first $k$ columns are the shared ones. 
\end{enumerate}
Here, $D^{(1)}_{i,p}$ ($D^{(2)}_{i,p}$) denotes the value in the $i$-th row and $p$-th column in dataset $D_1$ (respectively, dataset $D_2$). 

\subsection{Matrix Completion}

Given the merged data matrix representation $M$ for two allied, aligned datasets, the problem of constructing the unified dataset $D_{unified}$ can now be posed as the problem of completing the merged data matrix $M$ i.e., estimating the missing values denoted by $NA$. There are $N_1 \times n_2 + N_2 \times n_1$  missing entries in the matrix $M$, out of a total of $(N_1 + N_2) \times (k + n_1 + n_2)$ entries. 

Many methods have been devised for {\em matrix completion}~ \citep{PV25},  \citep{Jafa22},  \citep{RYLL18}. We used the {\em iterative SVD} method of matrix completion, which works well when the number of rows is much greater than the number of columns, to unify the two allied and aligned datasets i.e., to estimate the $NA$ values in the merged data matrix representation.

As an example, for OULAD and HEPSR allied datasets (after alignment), $N_1 = 32593, n_1 = 20, N_2 = 3630, n_2 = 28, k = 7$. The total number of entries in the merged data matrix is $(32593 + 3630) \times (7 + 20 + 28) = 1992265$, out of which the number of missing entries is $32593 \times 28 + 3630 \times 20 = 985204$ ($49.45$\%). Specifically, $n_2$ entries are missing in each of the first $N_1$ rows (for OULAD students) and $n_1$ entries are missing in each of the next $N_2$ rows after the first $N_1$ rows (for HEPSR students). These missing entries are estimated and the unified data matrix is created very efficiently by the iterative SVD matrix completion method. 

\section{Empirical Results}\label{sec_results}

\small
\begin{ThreePartTable}
\begin{TableNotes}
\small
 \item[1] When the test dataset is not specified, the $P, R, F_1$ numbers refer to $5$-fold cross-validation on training dataset. Moreover, they are for the minority class $1$.
 \item[2] \textbf{cleaned}: original rows and columns (after pre-processing and alignment)
 \item[3] \textbf{extended}: original rows and columns (after pre-processing and alignment) + extended columns 
 \item[3] \textbf{unified}: union of all rows + all extended columns
\end{TableNotes}

\begin{longtable}{|L{3cm}||L{2cm}||l|l|l||l||l||l|}
\caption{Classifier Performance on Allied Datasets in Various Scenarios} \label{tabexp} \\
\hline
\textbf{Training} & \textbf{Test} & \multicolumn{3}{c||}{\textbf{XGBoost}} &  \textbf{RF} &  \textbf{SVM} & \textbf{KNN}\\ \cline{3-5}
  & & \textbf{Prec.} & \textbf{Recall} & $\mathbf{F_1}$\tnote{1} & $\mathbf{F_1}$ & $\mathbf{F_1}$ & $\mathbf{F_1}$ \\
\hline
\endfirsthead

\multicolumn{8}{l}{\textit{Table~\ref{tabexp}  continued from previous page}} \\
\hline
\textbf{Training} & \textbf{Test} & \multicolumn{3}{c||}{\textbf{XGBoost}} &  \textbf{RF} &  \textbf{SVM} & \textbf{KNN}\\ \cline{3-5}
  & & \textbf{Prec.} & \textbf{Recall} & $\mathbf{F_1}$ & $\mathbf{F_1}$ & $\mathbf{F_1}$ & $\mathbf{F_1}$ \\
\hline
\endhead

\hline
\insertTableNotes \\ 
\endlastfoot
\footnotesize
oulad\_ cleaned\tnote{2} & $-$ & 0.7535 & 0.7535 & \textcolor{red}{0.7783} & \textcolor{red}{0.7761} & \textcolor{red}{0.7938} &  \textcolor{red}{0.7351}\\
\hdashline
hepsr\_ cleaned & $-$ & 0.9136 & 0.9136 & \textcolor{red}{0.8741} & \textcolor{red}{0.8684} & \textcolor{red}{0.8730} &  \textcolor{red}{0.7641}\\
\hdashline
oulad\_ extended\tnote{3} & $-$ & 0.9541 & 0.9541 & \textcolor{blue}{0.9516} & \textcolor{blue}{0.8687} & \textcolor{blue}{0.9929} & \textcolor{blue}{0.8165}\\
\hdashline
hepsr\_  extended & $-$ & 1.0000 & 1.0000 & \textcolor{blue}{0.9989} & \textcolor{blue}{0.9996} & \textcolor{blue}{0.9951} & \textcolor{blue}{0.9822}\\
\hdashline
student\_dropout\_ unified\tnote{4} & $-$ & 0.9610 & 0.9610 & 0.9568 & 0.8829 & 0.9923 & 0.8244\\
\hdashline
student\_dropout\_ unified & oulad\_ extended & 1.0000 & 0.9999 & \textbf{1.0000} & \textbf{1.0000} & \textbf{0.9971} & \textbf{1.0000}\\
\hdashline
student\_dropout\_ unified & hepsr\_ extended & 1.0000 & 1.0000 & \textbf{1.0000} & \textbf{1.0000} & \textbf{1.0000} & \textbf{1.0000}\\
\hdashline
oulad\_ extended & hepsr\_ extended & 0.8798 & 0.7108 & 0.7863 & 0.8102 & 0.0000 & 0.6408\\
\hdashline
hepsr\_ extended & oulad\_ extended & 0.5286 & 0.9596 & 0.6817 & 0.7146 & 0.7772 & 0.7102\\
\hline\hline
oracle\_ cleaned & $-$ & 0.3468 & 0.3468 & \textcolor{red}{0.1129} & \textcolor{red}{0.0022} & \textcolor{red}{0.2283} &  \textcolor{red}{0.0507}\\
\hdashline
mendeley\_ cleaned & $-$ & 0.5568 & 0.5568 & \textcolor{red}{0.4719} & \textcolor{red}{0.1348} & \textcolor{red}{0.3457} &  \textcolor{red}{0.1502}\\
\hdashline
oracle\_ extended & $-$ & 0.9989 & 0.9989 & \textcolor{blue}{0.9715} & \textcolor{blue}{0.8463} & \textcolor{blue}{0.9901} & \textcolor{blue}{0.7479}\\
\hdashline
mendeley\_ extended & $-$ & 0.9915 & 0.9915 & \textcolor{blue}{0.9644} & \textcolor{blue}{0.8718} & \textcolor{blue}{1.0000} & \textcolor{blue}{0.3881}\\
\hdashline
auto\_fraud\_ unified & $-$ & 0.9847 & 0.9847 & 0.9541 & 0.7814 & 0.8735 & 0.3441\\
\hdashline
auto\_fraud\_ unified & oracle\_ extended & 1.0000 & 1.0000 & \textbf{1.0000} & \textbf{1.0000} & \textbf{1.0000} & \textbf{1.0000}\\
\hdashline
auto\_fraud\_ unified & mendeley\_ extended & 1.0000 & 1.0000 & \textbf{1.0000} & \textbf{1.0000} & \textbf{1.0000} & \textbf{1.0000}\\
\hdashline
oracle\_ extended & mendeley\_  extended & 0.3083 & 0.1660 & 0.2158 & 0.3203 & 0.0000 & 0.2081\\
\hdashline 
mendeley\_ extended & oracle\_ extended & 0.1118 & 0.1170 & 0.1143 & 0.0878 & 0.0021 & 0.0160\\
\hline\hline
umair\_ cleaned & $-$ & 1.0000 & 1.0000 & \textcolor{red}{1.0000} & \textcolor{red}{1.0000} & \textcolor{red}{0.9925} &  \textcolor{red}{0.8735}\\
\hdashline
sharma\_ cleaned & $-$ & 0.9850 & 0.9850 & \textcolor{red}{0.9801} & \textcolor{red}{0.9724} & \textcolor{red}{0.9161} &  \textcolor{red}{0.8601}\\
\hdashline
umair\_ extended & $-$ & 1.0000 & 1.0000 & \textcolor{blue}{1.0000} & \textcolor{blue}{1.0000} & \textcolor{blue}{0.9988} & \textcolor{blue}{0.8936}\\
\hdashline
sharma\_ extended & $-$ & 1.0000 & 1.0000 & \textcolor{blue}{0.9988} & \textcolor{blue}{0.9988} & \textcolor{blue}{1.0000} & \textcolor{blue}{0.9975}\\
\hdashline
loan\_approval\_ unified & $-$ & 1.0000 & 1.0000 & 0.9997 & 0.9996 & 0.9977 & 0.8944\\
\hdashline
loan\_approval\_ unified & umair\_ extended & 1.0000 & 1.0000 & \textbf{1.0000} &  \textbf{1.0000} & \textbf{0.9993} & \textbf{1.0000}\\
\hdashline
loan\_approval\_ unified & sharma\_ extended & 1.0000 & 1.0000 & \textbf{1.0000} & \textbf{1.0000} & \textbf{0.9975} & \textbf{1.0000}\\
\hdashline
umair\_ extended & sharma\_ extended & 1.0000 & 0.3943 & 0.5656 & 0.5611 & 0.5495 & 0.5564\\
\hdashline
sharma\_ extended & umair\_ extended & 0.8971 & 0.5584 & 0.6884 & 0.7533 & 0.0000 & 0.6276\\
\hline\hline
saquib\_ cleaned & $-$ & 0.1614 & 0.1614 & \textcolor{red}{0.0409} & \textcolor{red}{0.0067} & \textcolor{red}{0.1592} &  \textcolor{red}{0.0288}\\
\hdashline
iiot\_ cleaned & $-$ & 0.4169 & 0.4169 & \textcolor{red}{0.2664} &  \textcolor{red}{0.0971} & \textcolor{red}{0.3399} &  \textcolor{red}{0.1069}\\
\hdashline
saquib\_ extended & $-$ & 1.0000 & 1.0000 & \textcolor{blue}{0.9983} & \textcolor{blue}{1.0000} & \textcolor{blue}{1.0000} & \textcolor{blue}{1.0000}\\
\hdashline
iiot\_ extended & $-$ & 0.4169 & 0.4169 & \textcolor{blue}{0.2664} & \textcolor{blue}{0.097}1 & \textcolor{blue}{0.3399} & \textcolor{blue}{0.1069}\\
\hdashline
machine\_failure\_ unified & $-$ & 0.8816 & 0.8816 & 0.7867 & 0.7915 & 0.7772 & 0.7744\\
\hdashline
machine\_failure\_ unified & saquib\_ extended & 1.0000 & 1.0000 & \textbf{1.0000} &  \textbf{1.0000} & \textbf{1.0000} & \textbf{1.0000}\\
\hdashline
machine\_failure\_ unified & iiot\_ extended & 1.0000 & 1.0000 & \textbf{1.0000} & \textbf{1.0000} & 0.9501 & \textbf{1.0000}\\
\hdashline
saquib\_ extended & iiot\_ extended & 0.2466 & 0.3395 & 0.2857 & 0.3320 & 0.0000 & 0.3619\\
\hdashline
iiot\_ extended & saquib\_ extended & 0.1571 & 0.1745 & 0.1653 & 0.0192 & 0.3445 & 0.1611\\
\hline\hline
\end{longtable}
\end{ThreePartTable}
\normalsize

\begin{figure}[htbp]
\includegraphics[width=0.9\columnwidth]{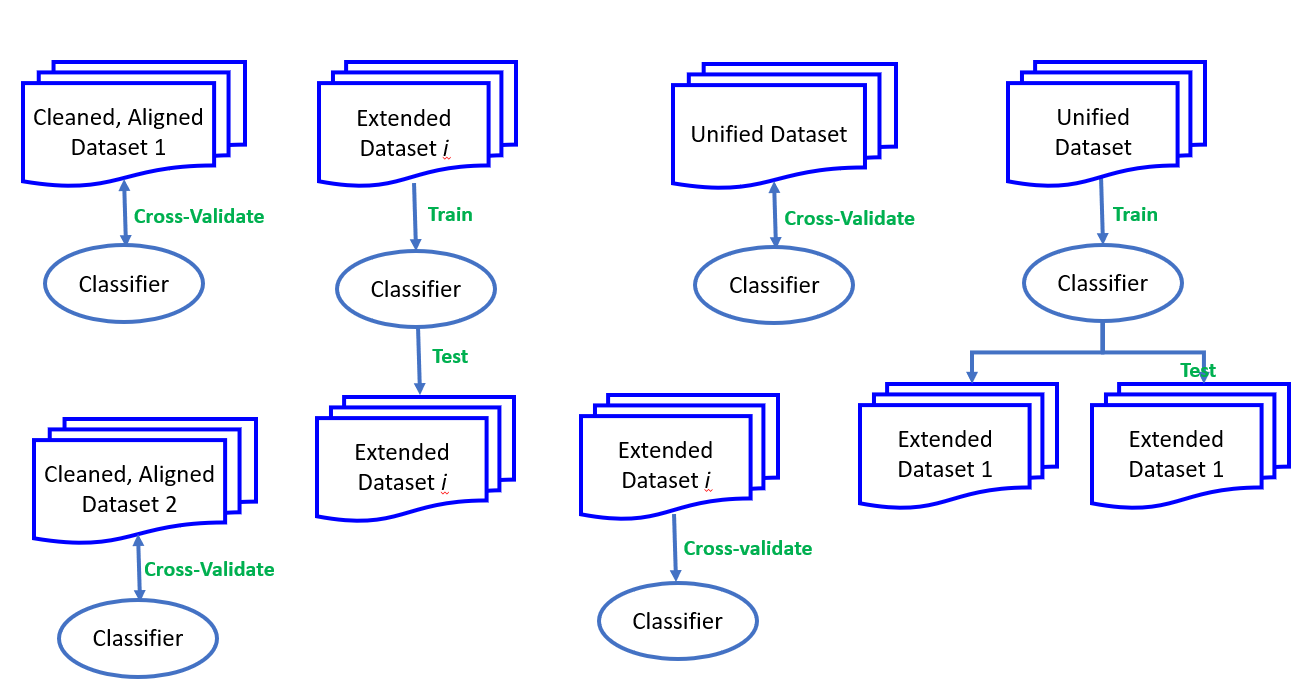}
\caption{Nine different experimental scenarios.}
\label{fig_exp_scenarios}
\end{figure}

For each pair of allied, and aligned datasets (Table~\ref{tab_ex}), we performed experiments under several scenarios (Fig.~\ref{fig_exp_scenarios}), where we trained several classical ML models (classifiers) and computed the associated performance measures such as Precision, Recall and $F_1$. We did not do any feature engineering, except standardizing all numeric features. We have also not performed any hyper-parameter tuning, and used only the default values for the hyper-parameters (e.g., RBF kernel and $C = 1$ for{\small {\sf SVM}}, $k = 5$ for {\small {\sf KNN}} etc.). Table~\ref{tabexp} shows the results for four classifiers {\small {\sf XGBoost, RandomForest, SVM, KNN}}; results for other classifiers were similar. We used the classifier implementations in Python's {\small {\tt scikit-learn}} package. Using classifier implementations in {\small {\tt pyCaret}} package gave similar results. Since both the datasets $D_1$ and $D_2$ now have the {\em same} feature-space in the cleaned, extended and unified versions, we are free to try various combinations, such as (i) train a classifier on one extended dataset and test the {\em same} classifier on the other extended dataset, or (ii) train a classifier on the unified dataset and test it on {\em each} constituent extended dataset, as shown in the Table. Broad observations about the results are as follows:

\begin{enumerate}
\item Numbers in \textcolor{red}{red} are the baseline $F_1$-measures (5-fold cross-validation) on the cleaned and aligned versions of the datasets. Numbers in \textcolor{blue}{blue} are the $F_1$-measures (5-fold cross-validation) on the extended versions, where more columns were added to each constituent dataset and whose values were estimated using matrix completion. The training and testing was only on the constituent dataset, but with extended columns. As seen, adding extended columns has clearly improved the performance of all classifiers for all allied datasets, as compared to the baselines (which used the original cleaned, aligned data but {\em without} extended columns). Thus we can unambiguously infer that adding extended column to one dataset, which were computed from the other, allied dataset is clearly beneficial.

For OULAD, the baseline $F_1$-measure on the cleaned, aligned dataset is $0.7783$ ($5$-fold cross-validation) for {\small {\sf XGBoost}}, whereas when using the cleaned, {\em and} extended OULAD dataset, the $F_1$ value for {\small {\sf XGBoost}} improves to $0.9516$ ($\Delta = 0.1733$, a clear $22.3$\% improvement). For HEPSR, the corresponding numbers are: $0.8741$ ({\small {\sf XGBoost}} on cleaned), $0.9989$ ({\small {\sf XGBoost}} on extended), $\Delta = 0.1248$ ($14.3$\% improvement). The $F_1$ and hence the improvement level is more for {\small {\sf RandomForest}} on extended OULAD dataset. The same trend is seen on all pairs of allied datasets and all classifiers that we have tried. {\em Only adding the extended columns to the original dataset improves the classifier cross-validation performance.} Since the values in the extended columns for $D_1$ are estimated using rows in {\em both} $D_1$ and $D_2$, this clearly shows that the knowledge from $D_2$ is transferred and learned by the classifier trained only on $D_1$ extended, even though the rows for $D_2$ are {\em not} used in the training on extended $D_1$. 
\item Training and testing on the entire unified dataset (which has rows from both datasets and all extended columns) yields very good performance. These numbers are always better than the baselines, though not necessarily the best. For example, training {\small {\sf XGBoost}} on the {\small {\sf auto\_fraud\_unified}} dataset, yields the 5-fold cross validation $F_1 = 0.9541$, much better than the baselines. Thus {\em training a classifier on the entire unified dataset yields very good cross-validation performance, better than the corresponding numbers on only the cleaned constituent datasets.}
\item As seen, for all the classifiers and all allied datasets, training on the unified dataset and testing on the extended subsets (corresponding to extended constituent datasets) yields perfect $F_1$ scores ($1.0$, or very close), shown in \textbf{bold}. This may be attributed perhaps to the classifiers memorizing the data, since the test dataset is a subset of the unified dataset on which the classifier was trained.  
\item Interestingly, if a classifier is trained on the entire {\em unified} dataset (rows from both datasets and all extended columns), and tested on constituent extended datasets, the performance always improves, but not necessarily yielding the most improvement. For instance, the {\small {\sf XGBoost}} classifier trained on the {\small {\sf auto\_fraud\_unified}} dataset shows $F_1 = 0.8872$ and $F_1 = 0.8872$ respectively when tested on {\small {\sf oracle\_extended}} and {\small {\sf mendeley\_extended}} datasets. While these numbers are much improved over the baseline figures of $0.0889$ and $0.4497$ on the two corresponding cleaned datasets, they are less than the best numbers $0.9660$ (trained on {\small {\sf mendeley\_extended}} and tested on{\small {\sf oracle\_extended}}) and $0.9772$ (trained on {\small {\sf oracle\_extended}} and tested on{\small {\sf mendeley\_extended}}). This shows that the knowledge transfer is sometimes better from one constituent extended dataset to another, rather than from the unified dataset to constituent extended dataset. In either case, the extended columns are playing a critical role as carriers of knowledge, leading to substantial improvements. 
\item In the {\small {\sf loan\_approval}} datasets, the cleaned versions already deliver an extremely high performance, even without the extended columns. However, adding the extended columns does lead to tiny improvements, and certainly does not  {\em reduce} the original perfoemance figures. This shows that adding extended columns is always likely to be beneficial. 
\end{enumerate}

For standalone original OULAD data, the performance of the {\small {\sf XGBoost}} (Extreme Gradient Boosting) classifier is $P = 0.7553, R = 0.8260, F_1 = 0.7889$, with $10$-fold cross-validation. For the {\small {\sf Random Forest}} classifier, the performance on the same data is $P = 0.7588, R = 07982, F_1 = 0.7778$. When using the full unified data (rows in OULAD as well as HEPSR data, along with all extended columns i.e., estimated values for all missing entries), the performance of the {\small {\sf XGBoost}} classifier jumps to $P = 0.9632, R = 0.9547, F_1 = 0.9589$ and for {\small {\sf Random Forest}} classifier it jumps to $P = 0.8765, R = 0.8717, F_1 = 0.8740$, showing a substantial improvement. Similar substantial jumps are observed in all classifiers that we tried. Clearly, the knowledge hidden in both the OULAD and HEPSR data has also been learnt simultaneously by the classifier trained on the unified data, since rows for both OULAD and HEPSR data are present in the unified data and the extended columns added for OULAD rows (HEPSR rows) explicitly model information derived from original HEPSR data (respectively, original OULAD data). 

\begin{figure*}[ht]
\subfigure[]{\includegraphics[width=0.45\textwidth]{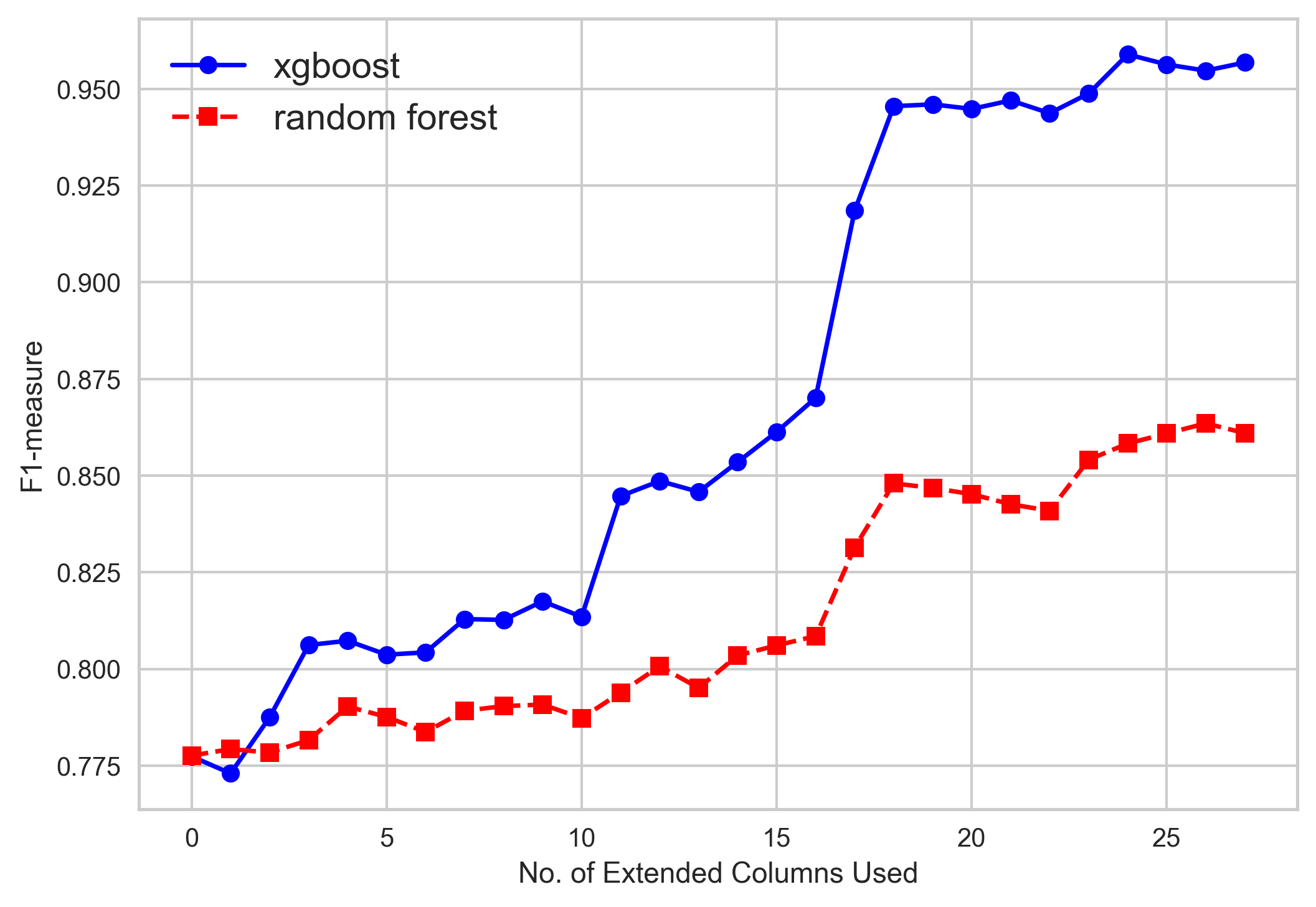}}
\subfigure[]{\includegraphics[width=0.45\textwidth]{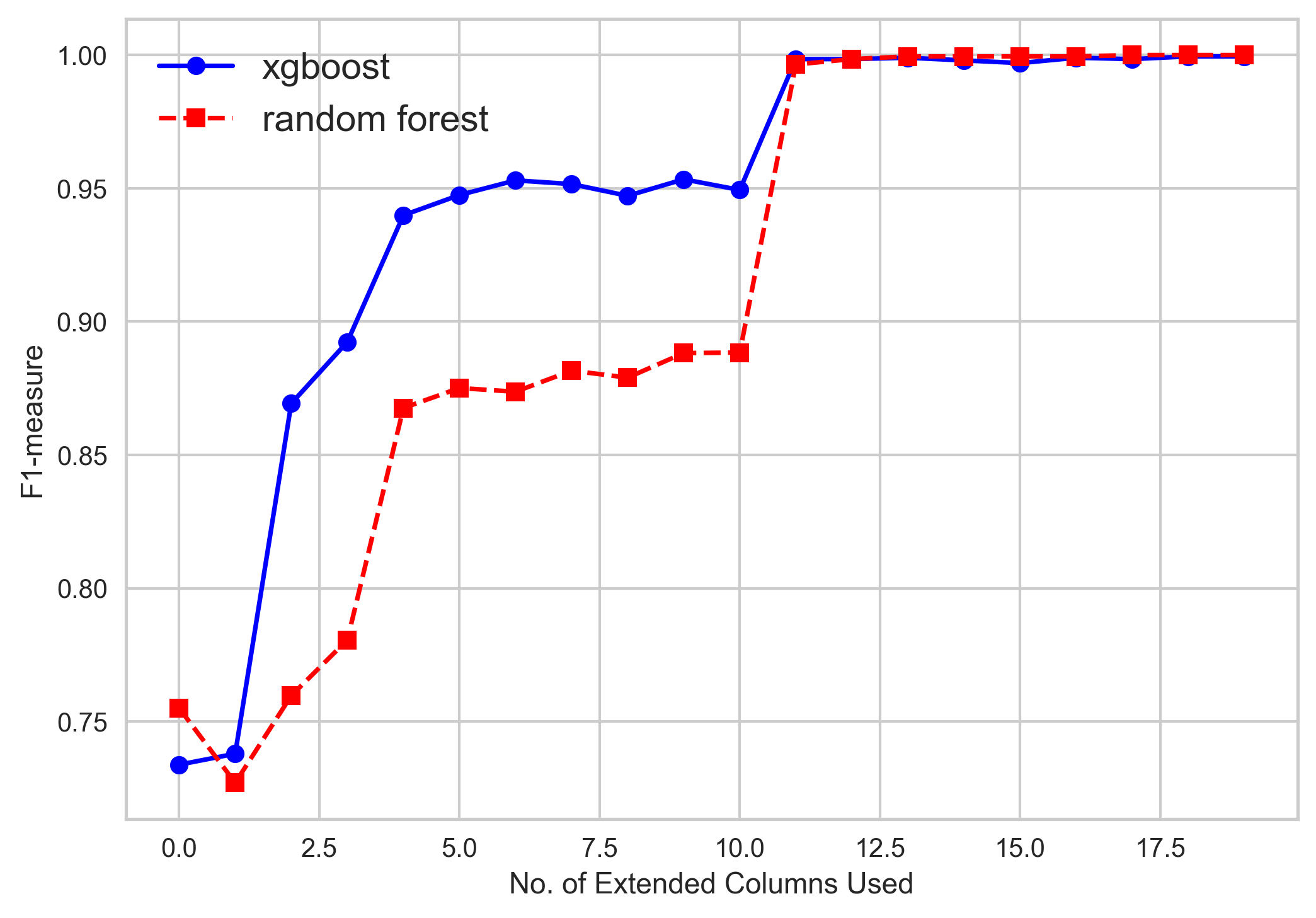}}
\caption{Change in $F_1$-measure vs. the number of extended columns used for (a) OULAD students (b) HEPSR students.}
\label{fig_F1}
\end{figure*}

Using only the first $N_1 = 32593$ rows corresponding to OULAD students, Fig.~\ref{fig_F1} shows the change in $F_1$ of two classifiers against the number of extended columns used. Clearly, for both classifiers, the performance keeps improving as more extended columns are used, even though the rows corresponding to HEPSR students are not used in the training. A similar improvement pattern is observed for HEPSR data. Note that the improvement for OULAD students is less for {\small {\sf RandomForest}} than {\small {\sf XGBoost}}, even when all extended columns are used. Moreover, as already shown, the performance of both classifiers improves on OULAD students, when using all extended columns and all rows (OULAD as well as HEPSR), demonstrating that the the classifier is learning from both sets of students. Dropping one set of students from training mildly reduces the size of the jump in the performance (for both classifiers). 

\begin{figure*}[ht]
\includegraphics[width=0.6\textwidth,height=6cm]{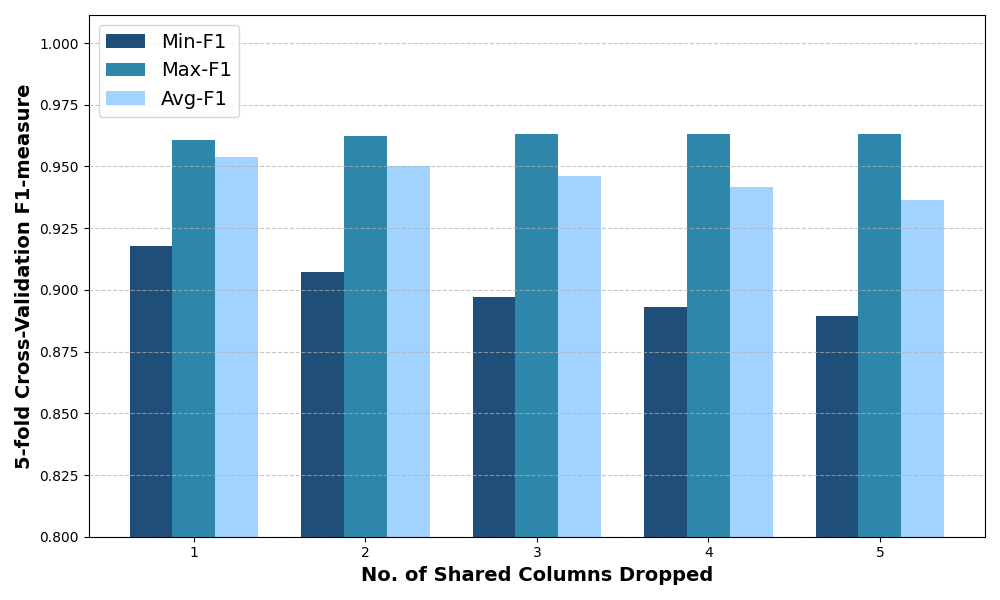}
\caption{Summary statistics for $F_1$-measure versus number of shared columns dropped from {\small {\sf auto\_fraud\_unified}} dataset .}
\label{fig_F2}
\end{figure*}

In another experiment, out of $k$ shared columns, we systematically dropped subsets of size $j$, $1 \leq j \leq k-1$ (column 0 is the class label) and computed the minimum, maximum and average $F_1$ for each unified dataset. Fig.~\ref{fig_F2} shows the result for {\small {\sf auto\_fraud\_unified}} dataset ($k = 13$). As seen, the minimum and average $F_1$ values (in 5-fold cross-validation) keep reducing as the number of dropped shared columns increases. Similar results are seen for other datasets and also when even more shared columns are dropped. Dropping all shared columns (other than the class label) yields the minimum, maximum and average $F_1$ as 0.0, 0.0 and 0.0 respectively. This shows that the shared columns have important influence on the classifier performance in the unified data. This makes sesnse because as the number of shared columns is reduced, the number of filled-up entries in the merged data matrix reduces and consequently, the matrix completion algorithm does a poorer job of estimating the missing data values. 

\section{Conclusions and Further Work}\label{sec_last}

In this paper, we defined the notion of {\em allied} datasets, which are labelled datasets with same class-labels, overlapping but otherwise distinct feature-spaces and related but distinct objects within the domain. We identified forur pairs of such allied datasets in diverse application domains: student dropout, insurance fraud, loan approval, and machine failures. In this paper, we have provided one way to directly transfer the classification knowledge from one dataset to another, or even unify such knowledge into a single whole. We proposed a method to merge the overlapping feature-spaces into a single feature-space for a pair of given allied heterogeneous datasets. We then used a matrix completion method to create a unified dataset based on the merged feature-space. We conducted experiments under different scenarios on these 4 pairs of allied, heterogeneous datasets and 4 different classifiers ({\small {\sf XGBoost, RandomForest, SVM, KNN}}). The experiments conclusively demonstrated that these classifiers, when trained on the unified representation, {\em always} outperformed classifiers, trained separately on the constituent allied datasets for {\em all} these pairs of allied datasets. 

This work provides an easy way to substantially improve classifier performance by unifying and using multiple allied datasets together. Since the datasets are independently created by different organizations on different domain objects, this approach allows to integrate them into a unified dataset, and build high-quality classification models that can be seamlessly applied to data in either format.  

We are working on applying the approach to several more pairs of allied datasets and also trying out more classifiers, including neural network models. We are also working on a GAN-like encoder-decoder based approach that maps the two distinct feature-spaces to a common latent space. The approach in this paper can be easily extended to unify, say, 3 allied datasets; the only constraint is that the shared attributes must be present in all 3 allied datasets. We are working on demonstrating the benefits of and evaluating any limitations of this larger data unification scheme. Another problem we are working on is evaluating the quality and interpretability of the newly added data values, estimated using various matrix completion algorithms i.e., theoretically modelling the data unification process (e.g., how does the performance vary per the number of shared columns?) and identifying true reasons for the substantially improved classification performance. One possibility is that the allied datasets share (hidden) reasons for producing the class label for a given data record, which the unified representatons are implicitly capturing through the extended columns. For example, almost certainly there are shared reasons why students drop-out in both OULAD and HEPSR datasets. It would be interesting to verify this hypothesis; currently the actual reasons as to why students drop out are not explicitly known in either dataset.

The current experiments show improvements in classifier performance under cross-validation. An interesting scenario would be to test the unified model - trained on unification of two allied datasets $D_1$ and $D_2$ - on a third ``unseen'' allied dataset $D_3$. For this, we are working out the scheme for estimating extended attributes based on $D_3$, without they being ``seen'' by the classifier trained on $D_1$ and $D_2$.

\bibliographystyle{elsarticle-num} 
\bibliography{ref1}  

\end{document}